\UseRawInputEncoding
\documentclass{article}
\PassOptionsToPackage{square, numbers}{natbib} %, compress
\PassOptionsToPackage{pdftex}{graphicx}
\usepackage[preprint]{neurips_data_2024}
\usepackage[utf8]{inputenc} % allow utf-8 input
\usepackage[T1]{fontenc}    % use 8-bit T1 fonts
\usepackage{hyperref}       % hyperlinks
\usepackage{url}            % simple URL typesetting
\usepackage{booktabs}       % professional-quality tables
\usepackage{amsfonts}       % blackboard math symbols
\usepackage{nicefrac}       % compact symbols for 1/2, etc.
\usepackage{microtype}      % microtypography
\usepackage{xcolor}         % colors
\usepackage[ruled,vlined]{algorithm2e}
\usepackage{graphicx}
\usepackage{soul}
\usepackage{tikz}
\usepackage{amssymb}

\usepackage{wrapfig}
\usepackage{caption}
\usepackage{mathtools}
\usepackage{multirow}
\usepackage{algpseudocode}
\usepackage{comment}
\usepackage{subcaption}
\usepackage{listings}
\usepackage{float}
\usepackage{enumitem}
\usepackage{listings}
\usepackage{xcolor}
\usepackage[pdftex]{graphicx}

\setlist[enumerate]{itemsep=0mm}
\setlist{nosep}

\newcommand{\mwp}{{\texttt{MWP-MISTAKE}}\xspace}
\newcommand{\gsm}{{\text{GSM-8K}}\xspace}
\newcommand{\mathd}{{\text{MATH}}\xspace}
\newcommand{\mathb}{{\text{MATHBENCH}}\xspace}
\newcommand{\jee}{{\text{JEEBENCH}}\xspace}
\newcommand{\gpt}{{\text{GPT-4}}\xspace}
\newcommand{\gptt}{{\text{GPT-3.5Turbo}}\xspace}
\newcommand{\gpto}{{\text{GPT-4o}}\xspace}

\newcommand{\claude}{{\text{Claude-3-Opus}}\xspace}
\newcommand{\PHI}{{\text{Phi-3-mini}}\xspace}
\newcommand{\llama}{{\text{Llama-2-7b-chat}}\xspace}
\newcommand{\mixtral}{{\text{Mixtral-8x7B}}\xspace}

\lstdefinestyle{promptstyle}{
  basicstyle=\ttfamily,
  breaklines=true,
  frame=single,
  captionpos=b,
  numbers=left,
  numberstyle=\tiny,
  numbersep=5pt,
}
\title{Exposing the Achilles' Heel: Evaluating LLMs Ability to Handle Mistakes in Mathematical Reasoning}

\author{%
Joykirat Singh \quad Akshay Nambi \quad Vibhav Vineet \\
Microsoft Research \\
\texttt{\{akshayn, vivineet\}@microsoft.com}\\
}

\begin{document}

\maketitle

\begin{abstract}
Large Language Models (LLMs) have been applied to Math Word Problems (MWPs) with transformative impacts, revolutionizing how these complex problems are approached and solved in various domains including educational settings. However, the evaluation of these models often prioritizes final accuracy, overlooking the crucial aspect of reasoning capabilities. This work addresses this gap by focusing on the ability of LLMs to detect and correct reasoning mistakes. We introduce a novel dataset \mwp, incorporating MWPs with both correct and incorrect reasoning steps generated through rule-based methods and smaller language models. Our comprehensive benchmarking reveals significant insights into the strengths and weaknesses of state-of-the-art models, such as \gpto, \gpt, \gptt, and others. We highlight \gpto's superior performance in mistake detection and rectification and the persistent challenges faced by smaller models. Additionally, we identify issues related to data contamination and memorization, impacting the reliability of LLMs in real-world applications. Our findings emphasize the importance of rigorous evaluation of reasoning processes and propose future directions to enhance the generalization and robustness of LLMs in mathematical problem-solving.

\end{abstract}

\section{Introduction}
\label{sec:intro}

Large Language Models (LLMs) have transformed artificial intelligence applications across diverse domains, including healthcare, agriculture, and education~\cite{noauthor_introducing_nodate_chat_gpt, noauthor_hello_nodate}. Their remarkable capabilities in natural language understanding, question answering, and mathematical problem-have shown potential to revolutionize various human endeavors~\cite{liu2024mathematical}. Recent advancements have fueled extensive research into applying LLMs to interpret and solve a wide array of mathematical tasks, from basic arithmetic to complex algebraic equations and calculus problems~\cite{hendrycks2021measuring, zhang2024careful}.

Math Word Problems (MWPs) convey mathematical concepts and calculations through written descriptions, typically involving narrative scenarios~\cite{srivatsa2024makes}. Solvers must extract relevant mathematical information from these narratives and apply appropriate principles to arrive at solutions. Studies~\cite{xu2024llms, heyueya2023solving, deb2023blank} have demonstrated that LLMs are proficient at understanding the contextual subtleties of MWPs, translating textual descriptions into mathematical expressions, and delivering precise solutions. Central to this process is \textit{mathematical reasoning}, which enables models to adeptly manage complex, multi-step problems, draw logical inferences, and provide accurate solutions.

Despite achieving remarkable accuracy rates exceeding 90\% on datasets like \gsm (Grade School Math dataset with linguistically diverse word problems)~\cite{cobbe2021training}, foundational LLMs such as \claude~\cite{noauthor_introducing_nodate}, Gemini Ultra~\cite{geminiteam2024gemini}, and OpenAI \gpt~\cite{openai2024gpt4} reveal a significant gap in our understanding of their capabilities in mathematical reasoning~\cite{deb2023blank}. Current research predominantly focuses on evaluating the final accuracy of MWPs~\cite{luo_wizardmath_2023, yu_metamath_2024}, neglecting the intricate reasoning processes necessary to derive solutions. We argue that the reasoning steps play a pivotal role, and it is imperative to assess them to comprehensively analyze the foundational capabilities of these models. This necessity is further underscored by the increasing utilization of LLMs in domains such as education~\cite{gan2023large}, where they serve as personalized tutors for students, aiding in teaching concepts and solving mathematical problems. Simply deriving the final answer is insufficient; the ability to guide students through correct steps, identify errors in their reasoning, and provide corrective guidance is paramount for such applications.
\begin{figure}[h!]
% \vspace{-20pt}
    \centering
    \includegraphics[width=\textwidth]{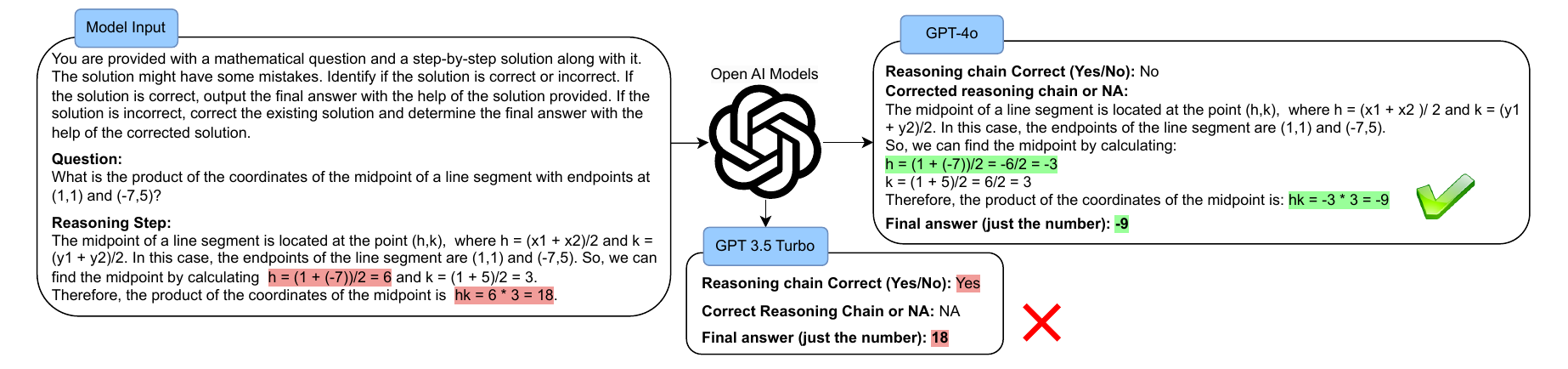}
    \caption{Model is prompted with a question along with \textcolor{red}{incorrect reasoning steps} to detect any mistake and correct the \textcolor{green}{reasoning step} to get to the correct final answer. \gpto generates the correct output, while \gptt fails to identify any mistake in the reasoning step. (Task - T1)}
    \label{fig:input_output}
    % \vspace{-20pt}
\end{figure}

This paper aims to bridge this gap by providing a comprehensive benchmark and evaluation of LLMs' performance on math word problems, including their capabilities in mistake detection and correction within the reasoning steps (Figure~\ref{fig:input_output}). Analyzing LLMs' ability to detect and rectify errors along the reasoning steps yields valuable insights into their overall problem-solving capabilities. Our objectives are threefold: firstly, to comprehensively evaluate LLMs' capabilities in mathematical reasoning, with a particular emphasis on mistake detection and correction; secondly, to identify the specific strengths and weaknesses of these models in handling various types of mathematical challenges; and thirdly, to propose potential directions for enhancing LLM capabilities in this domain.

To achieve this comprehensive evaluation, we have developed our own mistake dataset, designed to include errors in the reasoning steps. This dataset allows the assessment of models' proficiency not only in providing correct solutions but also in detecting and correcting mistakes within the reasoning steps. 
% We evaluate a total of models, including proprietary LLMs such as \gpt~\cite{openai2024gpt4}, \gpto~\cite{noauthor_hello_nodate}, \gptt~\cite{noauthor_openai_nodate}, and \claude~\cite{noauthor_introducing_nodate}, as well as smaller language models like \PHI~\cite{abdin2024phi3}, \llama~\cite{touvron2023llama}, and \mixtral~\cite{jiang2024mixtral}, using our curated dataset \mwp.
We evaluate eight different models including both large and smaller language models on our curated dataset \mwp.

Our analysis reveals several key insights into the performance of LLMs on MWPs. Firstly, detecting mistakes, even trivial ones remains a significant challenge for these models. Secondly, LLMs often derive correct answers despite this difficulty in mistake detection. This can be attributed to data memorization and potential contamination in training datasets, where models may have encountered similar/same problems before. However, the ability to recover from or correct errors in the reasoning process is generally poor across most models. 
% This indicates a gap in the LLMs' understanding and application of mathematical reasoning, highlighting the need for improved methodologies to enhance their problem-solving capabilities.
Our contributions to this paper are as follows:
\begin{enumerate}[leftmargin=*]
\itemsep0em 
   \item We collect and release to the research community \mwp, a dataset containing MWPs with both correct and incorrect reasoning obtained from state-of-the-art MWP datasets such as \gsm~\cite{cobbe_training_2021}, \mathd~\cite{hendrycks2021measuring}, \mathb~\cite{liu2024mathbench}, and \jee~\cite{arora2023llms}. Incorrect reasoning is derived through meticulously crafted rules to alter the reasoning steps and using smaller models, leveraging their inherent limitations in solving MWPs.

    \item We provide benchmark results for our dataset to evaluate the reasoning capabilities of state-of-the-art LLMs such as GPT-4o~\cite{noauthor_hello_nodate}, GPT-4~\cite{openai2024gpt4}, GPT-3.5Turbo~\cite{noauthor_openai_nodate}, Claude~\cite{noauthor_introducing_nodate}, as well as smaller language models like Llama~\cite{touvron2023llama}, Phi~\cite{abdin2024phi3}, and Mixtral~\cite{jiang2024mixtral}. Our analysis demonstrates that most state-of-the-art LLMs, excluding GPT-4o, struggle with mistake detection and correction.

    \item Through meticulous evaluation and comparison of different LLMs, we offer a detailed analysis of their strengths and weaknesses in handling mathematical reasoning tasks.
\end{enumerate}

\section{MWP-Mistake Dataset}
\label{sec:mwp-dataset}
\begin{figure}[h!]
% \vspace{-15pt}
    \centering
    \includegraphics[width=\textwidth]{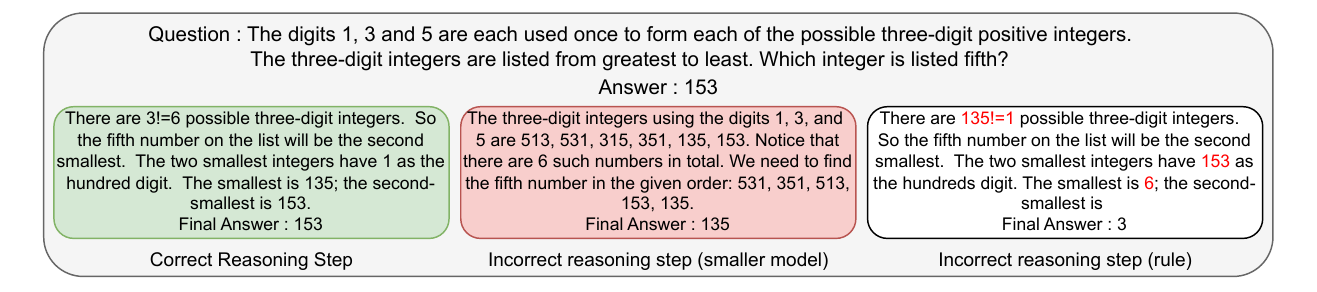}
    \caption{Examples of MWPs with correct reasoning, rule-based incorrect and smaller model based incorrect reasoning from \mathd.}
    \label{fig:examples}
    % \vspace{-15pt}
\end{figure}

% \vspace{-5pt}
Most MWP datasets include a math problem and the final answer, with some optionally providing reasoning steps (i.e., steps to solve the math problem) (See Figure.~\ref{fig:examples}). Our objective in this work is to evaluate the LLMs' ability to detect and rectify errors to derive the correct final answer. However, no existing datasets include incorrect reasoning steps for MWPs. To address this, we curated our own dataset, \mwp, by leveraging state-of-the-art MWP datasets such as \gsm~\cite{cobbe_training_2021}, \mathd~\cite{hendrycks2021measuring}, \mathb~\cite{liu2024mathbench}, and \jee~\cite{arora2023llms}. \mathb and \jee are relatively newer datasets as compared to \gsm and \mathd (Additional details in Appendix~\ref{app:dataset}).
% We now briefly discuss the four datasets we employ and then provide details on how we curate incorrect reasoning steps: 

% \begin{itemize}[leftmargin=20pt]
% \itemsep0em 
%     \item \gsm~\cite{cobbe_training_2021}:\gsm is a dataset of diverse grade school math word problems created by human writers, involving basic arithmetic operations. Released in November 2021.
%     \item \mathd~\cite{hendrycks2021measuring}: The \mathd dataset is divided into seven categories, each with five difficulty levels. For our study, we used levels 1, 2, and 3 from the algebra and counting and probability categories. Released in November 2021.
%     \item \mathb~\cite{liu2024mathbench}: \mathb is a recent dataset with questions divided by educational stages, from basic arithmetic to college levels. For our experiment, we chose middle and high school level single-choice multiple-choice questions. Released in May 2024.
%     \item \jee~\cite{arora2023llms}: \jee is a challenging benchmark dataset for evaluating LLM problem-solving abilities, containing 515 pre-engineering math, physics, and chemistry problems from the IIT JEE-Advanced Exam. For our experiment, we chose mathematics single-choice questions. Released in October 2023.
% \end{itemize}
Each dataset contains an MWP question and a final solution. While \gsm and \mathd have ground truth reasoning steps, \mathb and \jee do not. For these datasets, we used \gpt to curate chain-of-thought reasoning steps. Thus, for all four datasets, we have an MWP question, a final answer, and associated correct reasoning steps. Also, note that in \gsm and \mathd, the reasoning steps might include the final answer, however in our COT-generated steps, we ensure the answer is not present in the reasoning steps (Appendix~\ref{app:dataset} for additional details). 

To create incorrect reasoning steps, we follow two approaches: (i) meticulously crafted rules, and (ii) using smaller models as bad reasoners, which we describe next. 
% \vspace{-6pt}
\subsection{Meticulously Crafted Rules to Programmatically Inject Errors}
\label{sec:rules_errors}
% \vspace{-10pt}
Given our focus on MWPs and based on extensive interactions with math teachers, the rules are derived from common mistakes observed in educational settings, ensuring the errors introduced are realistic and representative of actual student errors.

\begin{enumerate}[leftmargin=*]
\item \textbf{Shuffle reasoning steps:} The reasoning steps are shuffled to introduce ambiguity in the thought process. This tests whether the model can identify changes in reasoning order.
\item \textbf{Delete reasoning steps:} One reasoning step is deleted in solutions that have two or more steps. This helps to identify if the model can spot omissions in the reasoning process.
\item \textbf{Shuffle numerical values:} Numerical values are shuffled among themselves to verify if models can correctly understand the question and select appropriate numerical values from the question.
\item \textbf{Replace numerical values:} Numerical values are replaced with random numbers ranging from 0 to 100. It identifies if the model can correctly pick the numerical values present in the question.
\item \textbf{Shuffle operations:} We randomly swap operators with other operators to test the model's ability to perform numerical operations.
\item \textbf{Insert random reasoning steps:} A random reasoning step is added at a random position to test the model's ability to identify incorrect reasoning.
\end{enumerate}
\begin{wraptable}{r}{0.72\textwidth}

% \begin{table}[]
\centering
\caption{\mwp Dataset details with the total number of questions and reasoning steps. }
\label{tab:dataset}
\resizebox{0.75\textwidth}{!}{%
\begin{tabular}{|l|ll|lll|l|}
\hline
\multicolumn{1}{|c|}{\multirow{3}{*}{Dataset}} & \multicolumn{2}{c|}{Default reasoning} & \multicolumn{3}{c|}{Smaller model reasoning} &      \\ \cline{2-7} 
\multicolumn{1}{|c|}{} &
  \multicolumn{1}{l|}{\multirow{2}{*}{\begin{tabular}[c]{@{}l@{}}\# Questions\\ with correct \\ reason (GT)\end{tabular}}} &
  \multirow{2}{*}{\begin{tabular}[c]{@{}l@{}}\# Questions\\ with incorrect \\ reason (Rules)\end{tabular}} &
  \multicolumn{3}{l|}{\begin{tabular}[c]{@{}l@{}}\# Questions\\ with incorrect reasoning\end{tabular}} &
  Total \\ \cline{4-7} 
\multicolumn{1}{|c|}{}                         & \multicolumn{1}{l|}{}        &         & \llama          & \mixtral          & \PHI          &      \\
                                               & \multicolumn{1}{l|}{}        &         &                &              &              &      \\
\gsm                                          & \multicolumn{1}{l|}{93}      & 558     & 100            & 100          & 100          & 951  \\
\mathd                                           & \multicolumn{1}{l|}{150}     & 900     & 150            & 150          & 150          & 1500 \\
\mathb                                      & \multicolumn{1}{l|}{100}     & 600     & 100            & 100          & 100          & 1000 \\
\jee                                       & \multicolumn{1}{l|}{38}      & 228     & 12             & 19           & 35           & 332  \\ \hline
\end{tabular}%
}
% \vspace{-15pt}
% \end{table}
\end{wraptable}
These rules mimic real-world student behavior by reflecting tendencies to get the order of steps wrong, skip steps, misinterpret numerical values, use incorrect numbers, apply the wrong mathematical operations, and add irrelevant steps in problem-solving. While rules \#1 and \#2 do not introduce explicit errors in reasoning, they are considered mistakes in our dataset to prompt the model to identify scenarios lacking clarity. Such scenarios, whether due to an incorrect thought process or missing steps, are common in real-life situations. Table~\ref{tab:dataset} shows the number of questions selected from each of the four datasets to which these six rules are applied to curate incorrect reasoning. Thus, for every question selected, we created seven variations of reasoning steps(one correct + six incorrect).
% \usepackage{multirow}
% \usepackage{graphicx}

% \vspace{10pt}
\subsection{Smaller Models as Bad Reasoners}
\label{sec:small-error}
% \vspace{-9pt}
Recently, numerous small language models (SLMs) have been developed that are smaller in size and trained on smaller amounts of data. These models are highly efficient, effective, and capable of solving numerous tasks, including MWPs. However, they still lack several capabilities, including advanced mathematical reasoning, resulting in poorer performance on MWPs.

To curate incorrect reasoning steps, we use SLMs to generate Chain-of-Thought (COT) reasoning and final answers for all dataset questions. We filter out questions with incorrect final answers (comparing with the ground truth final answer from the dataset), assuming incorrect answers stem from incorrect reasoning. Thus, the reasoning steps for all incorrect answers are used as incorrect reasoning steps.
We employ state-of-the-art SLMs, such as \llama, \PHI, and \mixtral, to generate COT reasoning steps without a final answer (Appendix~\ref{app:slm_reasoning_steps} for additional details). Table~\ref{tab:dataset} shows dataset stats for each of the three models across all datasets.

Thus, our dataset includes questions with original correct reasoning steps, rule-based incorrect reasoning, and smaller model (SLM) generated incorrect reasoning. For detailed evaluation, we split this data into two parts: (1) \texttt{Default:} containing questions with correct reasoning from the dataset and rule-based incorrect reasoning, and (2) \texttt{SLM reason:} containing questions with incorrect steps generated by SLMs.
Table~\ref{tab:dataset} provides the complete details of the curated \mwp dataset with the above two splits. We are releasing this dataset for further evaluation and benchmarking. 
% \vspace{-10pt}
\section{Experimental Setup}
\label{sec:setup}
% \vspace{-4pt}
\textbf{Task Details.} Our aim is to assess the performance of LLMs on MWPs, focusing on their ability to detect and correct mistakes within the reasoning steps. We have two task variants to accomplish this:

\textbf{Task-1 (T1):} Here, given a question and its reasoning steps, we ask the model to identify if the steps are correct or incorrect. If incorrect, the model must rectify the mistake and calculate the final answer. The final answer or corrected reasoning step can be either correct or incorrect (Figure~\ref{fig:input_output}).   \\
\textbf{Task-2 (T2):} In this scenario, the model only needs to identify whether the reasoning steps provided are correct or incorrect and provide the final answer. No correction of reasoning steps is required. 

In essence, T1 evaluates the model's ability to detect mistakes, rectify them, and derive the correct answer, while T2 focuses solely on detecting mistakes and solving MWP correctly. Both tasks operate under few-shot settings, with specific prompt details provided in Appendix~\ref{app:t1_t2_details}.

\textbf{Models.} To evaluate LLMs' mathematical reasoning capabilities, we utilize state-of-the-art LLMs and Small Language Models (SLMs).

\textbf{LLMs:} We utilize LLMs that have shown tremendous performance in MWPs such as \gpto, \gpt, \gptt, \claude. These models are accessed through their respective APIs. \\
\textbf{SLMs.} Additionally, we assess SLMs trained with high-quality data and reasoning capabilities and explored three popular SLMs with diverse capabilities: \PHI, \mixtral, and \llama.
Appendix~\ref{app:model_details} provides the details of the models, including their last training date. 
% \vspace{-10pt}
\section{Results and Analysis}
\label{sec:results}
% \vspace{-5pt}
We rigorously evaluate various SOTA LLMs and SLMs on our \mwp dataset to analyze their mathematical reasoning capabilities, focusing on mistake detection and recovery.

% \usepackage{graphicx}
\begin{table}[h!]
\centering
\caption{Mistake Detection Performance (F1 score) on \mwp dataset for Task T1. (D-Default reasoning steps, SM-Smaller model reasoning steps) (Bold: Best, Underline:Second best)}
\label{tab:md_t1}
\resizebox{\textwidth}{!}{%
\begin{tabular}{|l|ll|ll|ll|ll|lll|}
\hline
                        & \multicolumn{2}{c|}{\gsm} & \multicolumn{2}{c|}{\mathd} & \multicolumn{2}{c|}{\mathb} & \multicolumn{2}{c|}{\jee} & \multicolumn{3}{c|}{Average}                  \\ \hline
Model                   & D                   & SM                 & D                    & SM                  & D                    & SM                  & D                   & SM                 & D             & SM            & Overall       \\ \hline
\gpto    & \textbf{0.85}       & \underline{0.84}         & \textbf{0.83}        & \textbf{0.86}       & \textbf{0.80}        & \textbf{0.99}       & \textbf{0.80}       & \textbf{0.99}      & \textbf{0.82} & \textbf{0.92} & \textbf{0.87} \\
\gpt     & 0.72                & 0.68               & 0.78                 & \underline{0.80}       & 0.51                 & 0.90                & \underline{ 0.74}          & 0.87               & 0.69          & 0.81          & 0.75          \\
\gptt    & \underline{ 0.80}          & 0.69               & \underline{ 0.80}           & 0.54                & 0.50                 & 0.34                & 0.54                & 0.46               & 0.66          & 0.51          & 0.58          \\
\llama   & 0.07                & NA                 & 0.16                 & NA                  & 0.08                 & NA                  & 0.41                & NA                 & 0.18          & NA            & 0.18          \\
\mixtral & 0.73                & NA                 & 0.79                 & NA                  & 0.62                 & NA                  & 0.70                & NA                 & 0.71          & NA            & 0.71          \\
\PHI     & 0.70                & NA                 & 0.65                 & NA                  & 0.54                 & NA                  & 0.67                & NA                 & 0.64          & NA            & 0.64          \\
\claude  & 0.79                & \textbf{0.87}      & 0.73                 & 0.76                & \underline{ 0.68}           & \underline{ 0.91}          & 0.69                & \underline{ 0.88}         & \underline{ 0.72}    & \underline{ 0.85}    & \underline{ 0.79}    \\ \hline
\end{tabular}
}
% \vspace{-12pt}
\end{table}
\vspace{-6pt}
% \vspace{-10pt}
\subsection{Question 1: Can LLMs Effectively Identify Mistakes in Reasoning Steps? }
\label{sec:results_md}
% \vspace{-4pt}
We first analyze the capability of various models to detect mistakes in MWP reasoning steps. Table~\ref{tab:md_t1} presents the mistake detection performance (F1 score) of all the models with Task T1 on our dataset, which includes reasoning steps derived from default and smaller models across four datasets. 

% \textbf{Performance Analysis.} The performance of \gpto significantly outshines other models, including \gpt, \gptt, smaller language models (SLMs), and \claude. 
\begin{itemize}[leftmargin=*]
\itemsep0em 
\item \textbf{\gpto's Dominance:} \gpto demonstrates a substantial advantage, with a 10\% improvement over \gpt, a 25\% improvement over \gptt, and over 20\% improvement over SLMs in detecting mistakes. It is uniquely capable of consistently identifying mistakes created using both rule-based methods and smaller models, underscoring its robust capabilities in mistake detection.

\item \textbf{\gptt's Performance:} Interestingly, \gptt outperforms \gpt in mistake detection specifically for the \gsm dataset. We hypothesize that this could be due to potential overfitting or data contamination in \gpt's training data. Despite this anomaly, \gpt maintains its position as the second-best model overall, following closely behind \gpto in terms of mistake detection abilities on other datasets. 

\item \textbf{Performance of SLMs:} 
SLMs show significantly lower mistake detection abilities compared to \gpto and \gpt. This stark contrast highlights the need to enhance reasoning capabilities in smaller models to match advanced LLMs.

\item \textbf{Performance on Newer Datasets:} The performance of most models, including \gpt and \gptt, drops drastically on newer datasets such as \mathb and \jee. This decline indicates that the reasoning abilities of these models are not yet generalized to newer datasets and problems. Furthermore, \jee is more challenging dataset compared to others. \gpto, however, maintains a significant lead even on these newer datasets, reinforcing its superior capability in mistake detection across diverse and unseen problems.
\end{itemize}

% Key Insighits and Takeaways:\\
% \textbf{1. GPT-4o's Superior Performance:} GPT-4o stands out as the most capable model in detecting reasoning mistakes, showcasing significant improvements over other state-of-the-art models.\\
% \textbf{2. Need for Improvement in SLMs:} The considerable gap in performance between SLMs and larger models like GPT-4 and GPT-4o emphasizes the necessity for advancements in the reasoning capabilities of smaller models.\\
% \textbf{3. Overfitting and Data Contamination Concerns:} The unexpected performance of GPT-3.5 Turbo over GPT-4 in certain datasets suggests issues related to overfitting and data contamination. We conduct additional experiments to quantify this in Section~\ref{}.\\
% \textbf{4. Generalization Challenges:} The notable performance drop on newer datasets highlights a critical challenge in the generalization of reasoning abilities in LLMs. 

Similar results are seen also for Task T2 as both T1 and T2 probes the model to detect mistakes, however, in the former case it goes further by asking the model to correct the reasoning step.

% \usepackage{graphicx}

\vspace{-6pt}
\subsection{Can LLMs Accurately Derive Correct Answers Despite Mistakes?}
\label{sec:results_perf}
\vspace{-6pt}
We now assess the models' ability to accurately derive the correct answer for the given question despite mistakes in the reasoning steps. Table~\ref{tab:perf_t1} shows the performance of all the models in deriving correct answers (F1 score) on our dataset.
\vspace{-10pt}
\begin{table}[h!]
\centering
% \vspace{-5pt}
\caption{Performance in deriving correct answers (F1 score) on \mwp dataset for Task T1. (D-Default reasoning steps, SM-Smaller model reasoning steps) (Bold: Best, Underline:Second best)}
\label{tab:perf_t1}
\resizebox{\textwidth}{!}{%
\begin{tabular}{|l|ll|ll|ll|ll|lll|}
\hline
                        & \multicolumn{2}{c|}{\gsm} & \multicolumn{2}{c|}{\mathd} & \multicolumn{2}{c|}{\mathb} & \multicolumn{2}{c|}{\jee} & \multicolumn{3}{c|}{Average}                  \\ \hline
Model                   & D                   & SM                 & D                    & SM                  & D                    & SM                  & D                   & SM                 & D             & SM            & Overall       \\ \hline
\gpto    & \textbf{0.99}       & \textbf{0.88}      & \textbf{0.90}        & \underline{0.79}          & \underline{0.90}           & \textbf{0.69}       & \underline{0.42}          & \textbf{0.47}      & \textbf{0.80} & \textbf{0.71} & \textbf{0.76} \\
\gpt     & 0.97                & 0.79               & 0.80                 & 0.69                & 0.88                 & 0.46                & 0.35                & \underline{0.27}         & 0.75          & 0.55          & 0.65          \\
\gptt    & 0.89                & 0.48               & 0.69                 & 0.35                & 0.75                 & 0.20                & 0.26                & 0.14               & 0.65          & 0.29          & 0.47          \\
\llama   & 0.80                & NA                 & 0.27                 & NA                  & 0.40                 & NA                  & 0.06                & NA                 & 0.38          & NA            & 0.38          \\
\mixtral & 0.87                & NA                 & 0.67                 & NA                  & 0.70                 & NA                  & 0.16                & NA                 & 0.60          & NA            & 0.60          \\
\PHI     & 0.88                & NA                 & 0.51                 & NA                  & 0.63                 & NA                  & 0.25                & NA                 & 0.57          & NA            & 0.57          \\
\claude  & \underline{0.98}          & \underline{0.88}         & \underline{0.89}           & \textbf{0.93}       & \textbf{0.92}        & \underline{0.51}          & \textbf{0.46}       & 0.26               & \underline{0.80}    & \underline{0.64}    & \underline{0.73}    \\ \hline
\end{tabular}
}
\end{table}

% \textbf{Performance Analysis.} The performance of \gpto significantly surpasses that of other models, even when there are mistakes in the reasoning steps.

% \paragraph{}
% \vspace*{-\parskip}
\begin{enumerate}[leftmargin=*]
\itemsep0em 
\item \textbf{\gpto's Superior Accuracy:} \gpto's ability to derive correct answers is notably higher LMs. \gpto outperforms \gpt by 10\%, \gptt Turbo by 30\%, and SLMs by a similar margin. We suspect the very high accuracy in \gsm may be due to data contamination, however on newer and complex datasets \gpto still outperforms other models. This indicates a strong capability to produce correct answers even when intermediate steps contain errors. 
\item \textbf{\gpt's Performance:} Interestingly, \gpt's ability to derive correct answers despite mistakes (F1 score of 0.97) is significantly better than \gptt Turbo (F1 score of 0.89). Yet, \gpt performs poorly in mistake detection (0.72 vs. 0.80). This improvement in deriving correct answers may potentially be due to data contamination, resulting in the memorization of problems in the \gsm dataset during \gpt's training.
\item \textbf{SLMs' Performance:} SLMs, particularly \mixtral, show performance very close to GPT-4 in deriving correct answers. This might again be due to its strong ability to produce correct answers in the presence of mistakes or data contamination during the training of SLMs, which allows them to recall correct answers despite reasoning mistakes.
\item \textbf{Performance on Newer and Complex Datasets:} On newer and more complex datasets such as \mathb and \jee, the performance significantly drops even for \gpto and more drastically for all other LLMs and SLMs. This highlights a critical limitation in the generalization of these models to newer and unseen problem sets.
% \item \textbf{Overall Competence of GPT-4o:} GPT-4o excels not only in detecting mistakes but also in deriving accurate answers despite those mistakes. This dual capability reinforces its superior performance across various types of mathematical problems and reasoning complexities.
\end{enumerate}
\begin{figure}[h!]

% \begin{wrapfigure}{l}{0.45\textwidth}
  \begin{center}
    \includegraphics[width=\textwidth]{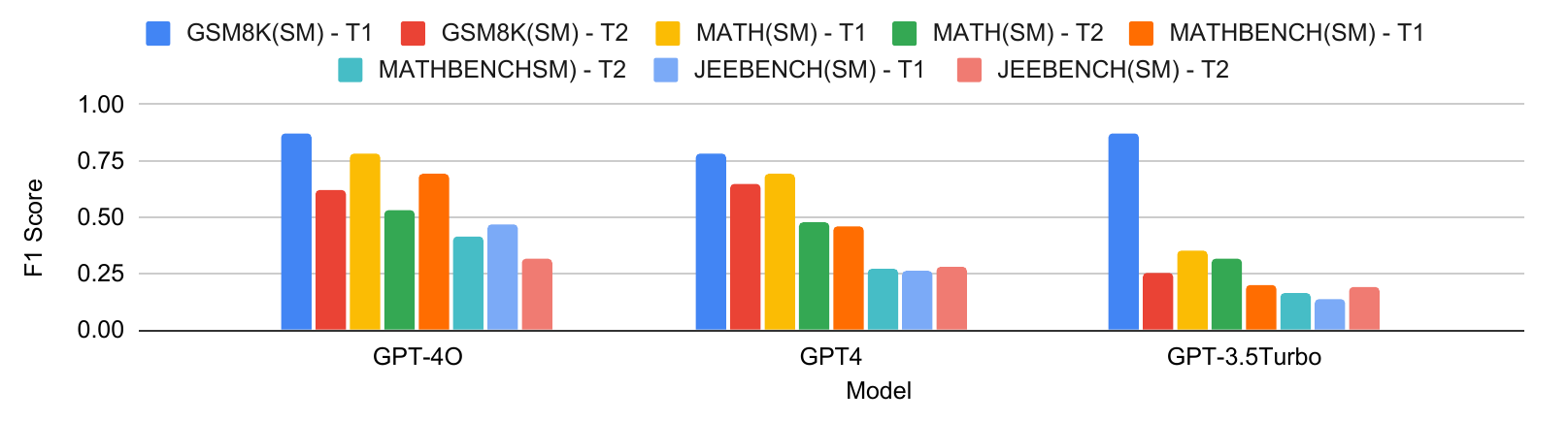}
  \end{center}
  % \vspace{-10pt}
  \caption{Performance in deriving final answer between T1 and T2. A significant drop in performance when the model does not rectify the incorrect reasoning steps.}
  % \vspace{-10pt}
  \label{fig:perf_t1_t2}
% \end{wrapfigure}
\end{figure}
Figure~\ref{fig:perf_t1_t2} shows the performance difference between T1 and T2. For T2, we observe a significant performance drop in deriving correct answers despite mistakes. This is primarily because, in T1, we instruct the model to not only detect mistakes but also correct them before deriving the final answer, whereas in T2, the model is only asked to detect the mistake and then directly derive the final answer without correcting the reasoning (Appendix~\ref{app:t2_details} for further details).

\subsection{Exploring Data Contamination and Memorization Effects in Math Reasoning Tasks}
\label{sec:mem}
% \vspace{-5pt}
In our analysis of LLMs' mathematical reasoning performance, we've identified potential instances of data contamination and memorization, both of which can significantly impact the effectiveness of these models. Data contamination, characterized by the presence of test data from downstream tasks in LLMs' training data, poses a major challenge in accurately assessing their real-world performance. Meanwhile, memorization occurs when models replicate solutions from training data without grasping the underlying principles, thereby hindering their ability to generalize to new problems. 

The presence of data contamination is evident in instances of unexpectedly high performance on certain datasets. For example, \gptt's superior performance over \gpt on the \gsm dataset raises concerns about biases in \gpt's training data. Similarly, the comparable performance between smaller and larger models suggests the potential presence of memorization. These findings underscore the critical need for rigorous evaluation to mitigate the impacts of memorization, ensuring the reliability and effectiveness of LLMs in real-world applications.

Investigating data contamination and memorization poses challenges due to restricted pre-training data access and computational limitations. To tackle this, we employ an approach outlined in~\cite{golchin2024time}, utilizing an LLM to replicate individual instances of the dataset. This involves guiding the LLM with instructions containing unique identifiers from the source dataset, like dataset name, partition (e.g., train, test, or validation), and a fragment of the reference instance. By instructing the LLM to complete these partial instances, we can evaluate contamination and memorization.

\begin{figure}[h!]
    \centering
    \includegraphics[width=\linewidth]{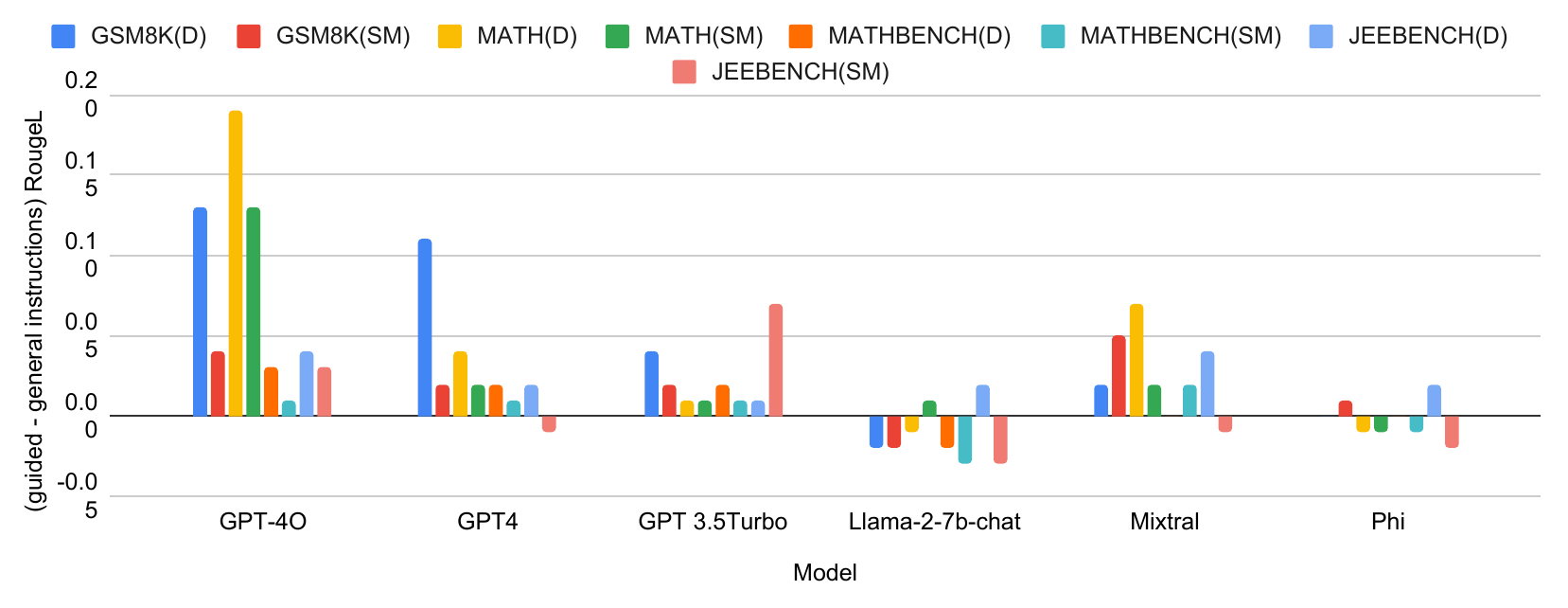}
    \caption{Difference between guided and general instructions rouge-L score across all models and datasets. A high positive difference indicates high contamination and a low positive or negative difference indicates, little to no contamination.}
    \label{fig:rouge_l}
    % \vspace{-20pt}
\end{figure}

To detect contamination, a heuristic is applied comparing the average overlap score between generated completions and reference instances using ROUGE-L~\cite{lin-2004-rouge}. This comparison is made between guided instructions (including dataset and partition identifiers) and general instructions (lacking such identifiers). If the overlap score is significantly larger with guided instructions, it suggests contamination. This method relies on the premise that the only distinction between the two instructions is the inclusion of dataset and partition names in guided instructions, implying any improvement can be attributed to contamination (Appendix~\ref{app:memo_prompt} for more details).
Figure~\ref{fig:rouge_l} shows the difference between guided and general instructions ROUGE-L score across all models and datasets. 

\begin{itemize}[leftmargin=*]
\itemsep0em
\item \textbf{\gpt Models:} Across all datasets for default reasoning steps, the guided scores are higher than the general scores, indicating contamination for all LLMs such as \gpto, \gpt, and \gptt.
\item \textbf{Smaller Models' Reasoning Mistakes:} For reasoning mistakes from smaller models (SM), guided scores are closer to general scores, indicating little to no contamination across all models. This is intuitive as the reasoning steps are created anew by smaller models, and due to their probabilistic nature, variations are expected.
\item \textbf{Smaller Models like \llama and \PHI:} These models show closer guided and general scores, indicating no contamination.
\item \textbf{\mixtral Model:} \mixtral shows greater contamination as compared to the rest of SLMs, explaining high performance when deriving correct answers.
\item \textbf{\gpto:} For datasets like \gsm and \mathd, \gpto shows a higher guided score than the general scores, indicating contamination, which decreases in contamination as the dataset becomes newer and more complex.

\end{itemize}

\vspace{-10pt}
\subsection{Can LLMs Correctly Rectify Mistakes in Reasoning Steps?}
\label{sec:results_rectify}

In Task 1, LLMs detect and rectify mistakes in reasoning to find the correct final answer. To evaluate the model's ability in this regard, we introduce the 'rectify metric' to quantify instances where the model identifies a mistake, corrects it, and reaches the accurate final answer. Reasoning steps are considered correct only if they lead to the accurate final answer. 
Table~\ref{tab:rect_t1} shows the ability of different models to rectify reasoning steps and derive the correct final answer across various datasets.

% \usepackage{graphicx}
\begin{table}[h!]
\centering
% \vspace{-10pt}
\caption{Ability to Rectify mistakes and derive correct final answer on \mwp dataset for Task T1. (D-Default reasoning steps, SM-Smaller model reasoning steps) (Bold: Best, Underline:Second best)}
\label{tab:rect_t1}
\resizebox{\textwidth}{!}{%
\begin{tabular}{|l|ll|ll|ll|ll|lll|}
\hline
                        & \multicolumn{2}{l|}{\gsm} & \multicolumn{2}{l|}{\mathd} & \multicolumn{2}{l|}{\mathb} & \multicolumn{2}{l|}{\jee} & \multicolumn{3}{l|}{Average}                  \\ \hline
Model                   & D                   & SM                 & D                    & SM                  & D                    & SM                  & D                   & SM                 & D             & SM            & Overall       \\ \hline
\gpto    & \textbf{0.98}       & \underline{0.92}         & \textbf{0.87}        & \underline{0.83}          & \textbf{0.90}        & \textbf{0.65}       & \textbf{0.39}       & \textbf{0.42}      & \textbf{0.79} & \textbf{0.70} & \textbf{0.74} \\
\gpt     & 0.96                & 0.89               & 0.72                 & 0.68                & 0.83                 & 0.46                & 0.23                & 0.24               & 0.69          & 0.57          & 0.63          \\
\gptt    & 0.81                & 0.58               & 0.54                 & 0.40                & 0.62                 & 0.35                & 0.05                & 0.05               & 0.51          & 0.35          & 0.43          \\
\llama   & 0.73                & NA                 & 0.21                 & NA                  & 0.11                 & NA                  & 0.04                & NA                 & 0.27          & NA            & 0.27          \\
\mixtral & 0.77                & NA                 & 0.56                 & NA                  & 0.57                 & NA                  & 0.17                & NA                 & 0.52          & NA            & 0.52          \\
\PHI     & 0.79                & NA                 & 0.37                 & NA                  & 0.41                 & NA                  & 0.03                & NA                 & 0.40          & NA            & 0.40          \\
\claude  & \underline{0.97}          & \textbf{0.94}      & \underline{0.84}           & \textbf{0.90}       & \underline{0.87}           & \underline{0.57}          & \underline{0.26}          & \underline{0.27}         & \underline{0.73}    & \underline{0.67}    & \underline{0.70}    \\ \hline
\end{tabular}
}
% \vspace{-10pt}
\end{table}

\begin{itemize}[leftmargin=*]
\itemsep0em   
    \item \textbf{\gpto's Remarkable Capabilities:} \gpto exhibits outstanding abilities in rectifying incorrect reasoning steps to derive the correct final answer. It outperforms \gpt by 11\% and surpasses other models, including SLMs, by over 35\%. Across all datasets, \gpto achieves high rectification scores, with an average of 85\% across all datasets except \jee. 
    \item \textbf{Limitations of SLMs:} SLMs perform notably worse than larger models like \gpto in rectifying errors, with an average score of only 40\% across all datasets. This suggests significant challenges in effectively handling complex reasoning tasks.
    \item \textbf{Performance on Newer and Complex Datasets:} Despite its overall superiority, \gpto's performance on newer and more complex datasets like \mathb and \jee is lower, raising concerns about the generalization of its capabilities. 
    \item \textbf{Ability to Rectify Mistakes from Both Rules and Smaller Models:} \gpto demonstrates tremendous capabilities in rectifying mistakes from both rule-based and smaller models. While potential contamination exists, \gpto’s ability to correct mistakes in the reasoning steps generated by SLMs underscores its robustness in detecting and rectifying errors.
\end{itemize}
% \begin{figure}[h!]
%     \centering
%     \includegraphics[width=\linewidth]{diff_model_output.pdf}
%     \caption{\gptt failed to identify any error in the \textcolor{red}{incorrect reasoning step}, leading to an incorrect final answer, whereas \gpto successfully detected the mistake and provided the \textcolor{green}{correct solution}.}
%     \label{fig:rouge_l}
%     % \vspace{-20pt}
% \end{figure}
We now dig deeper into the rectification process. While Table~\ref{tab:rect_t1} showed the models' ability to detect and rectify mistakes, we compute the percentage of questions where the model rectified the reasoning but still resulted in incorrect answers. Across the \mwp dataset, after correcting the reasoning steps, \gpto failed to derive correct answers in 17\% of the questions, whereas other models like \gpt, \gpt, \llama, \mixtral, and \PHI resulted in 30\%, 43.5\%, 80.9\%, 40.2\%, and 55.6\% incorrect answers, respectively. This showcases \gpto's ability to detect mistakes and rectify them correctly, resulting in very few questions it could not answer correctly.

Furthermore, we noticed that the average word length of rectified reasoning for correct and incorrect answers for \gpto was significantly higher than \gpt and other models. This is mainly because \gpto generates its own reasoning steps to rectify the mistakes, unlike other models that perform poorly. This also adds challenges to evaluating mistake rectification as the new rectified reasoning is significantly different from ground truth reasoning steps. There could be multiple ways to solve the same problems, complicating the evaluation.

We also evaluated the rectified reasoning steps and compared them with ground truth reasoning steps to see the effectiveness and alignment of the rectification process across models. We computed BERTScore~\cite{zhang2020bertscore} that computes a similarity score for each token in the candidate sentence with each token in the reference sentence, using BERT embeddings. 
% Table~\ref{tab:bert} shows the BERTScore derived for correct answers and incorrect answers obtained after rectification across all models and datasets. 
We found that BERTScore is similar across all models. This is because the BERTScore metric focuses on word-level matches and misses out on numerical and other logical aspects of reasoning which are crucial for correctness. We also evaluated the alignment with METEOR~\cite{banerjee-lavie-2005-meteor} score (see Appendix~\ref{app:meteor_results} for BERTScore and METEOR Score), which similarly resulted in an inadequate analysis. Thus, it becomes evident that the current evaluation methodologies may not fully capture the nuanced capabilities of LLMs in rectifying mistakes within reasoning steps.

\vspace{-10pt}
\section{Key Insights, Takeaways, and Potential Directions for Improving Mathematical Reasoning}
\label{sec:insights}
% \vspace{-5pt}
We now present an overview of key insights and takeaways obtained from our detailed benchmarking and evaluation of LLMs on our \mwp dataset. Further, we provide potential directions for improving mathematical reasoning abilities in LLMs. 

\begin{enumerate}[leftmargin=*]
\itemsep0em 

\item \textbf{\gpto's Superior Performance:} Despite potential data contamination, as observed in \gpto's performance, its superior foundational capabilities enable it to excel consistently across all datasets for mistake detection, rectification, and correct answer derivation. \gpto's remarkable performance positions it as a leading model for complex mathematical reasoning tasks, underscoring the robustness of its fundamental capabilities despite challenges such as data contamination.

\item \textbf{Challenges with SLMs:} The considerable performance gap between smaller language models (SLMs) and larger models like \gpt and \gpto emphasizes the necessity for advancements in the reasoning capabilities of smaller models. Enhancing these models could make them more competitive and useful in applications where resource constraints are significant.

\item \textbf{Overfitting and Data Contamination Concerns:} The unexpected performance of \gptt over \gpt in certain datasets suggests issues related to overfitting and data contamination. This is evident in the performance disparity, particularly in the \gsm dataset, indicating potential memorization of problems during training. Addressing these concerns requires cleaner training datasets and more robust methodologies to avoid overfitting and ensure genuine reasoning skills.

\item \textbf{Generalization Challenges:} The notable performance drop on newer datasets like \mathb and \jee underscores a critical challenge in generalizing LLMs' reasoning abilities to novel problems. Addressing this issue is crucial for enhancing the applicability and reliability of LLMs across a broader spectrum of mathematical problems and datasets.

\item \textbf{SLMs' Unexpected Performance:} The close performance of some SLMs, like \mixtral, to larger models such as \gpt suggests that these smaller models might also benefit from data contamination. This indicates a need for further investigation into training processes and dataset integrity to ensure fair and accurate performance assessments.
\end{enumerate}

These insights underscore the ongoing necessity to refine LLM training processes, enhance reasoning capabilities, and improve generalization to ensure models can reliably and accurately solve a wide range of mathematical problems. Future research should prioritize addressing overfitting, data contamination, and generalization challenges to advance LLMs in the field of mathematical reasoning.

% Future Directions
% To enhance the mathematical reasoning capabilities of LLMs, future research should focus on:

% Improving Mistake Detection: Developing techniques to better identify subtle and complex errors in reasoning steps.
% Enhancing Error Correction: Creating models or training methods that improve the ability to correct mistakes, thus providing more reliable step-by-step guidance.
% Robustness Against Data Memorization: Ensuring models understand reasoning processes rather than relying on memorized solutions from training data.

\vspace{-5pt}
\section{Related Work}
\label{sec:rw}
\vspace{-5pt}
Recent studies \cite{wang2023selfconsistency} indicate that Large Language Models (LLMs) can handle intricate tasks using the Chain of Thought (COT) mechanism \cite{wei2023chainofthought}. LLMs have gained significance in solving math word problems (MWPs) \cite{liu2024mathematical}, with MathPrompter \cite{imani2023mathprompter} showcasing excellent results, not only generating correct answers but also complex reasoning steps. Various approaches aim to enhance LLMs' mathematical capabilities and address challenges \cite{srivatsa2024makes}. \cite{yuan2023scaling} investigates factors like pre-training loss, supervised data, and augmented data, proposing rejection sampling fine-tuning (RFT) to improve mathematical reasoning. WizardMath \cite{luo2023wizardmath} introduces a reinforced Evol-Instruct Feedback (RLEIF) method to enhance reasoning abilities through supervised fine-tuning and PPO training \cite{schulman2017proximal}. MAmmoTH \cite{yue2023mammoth} combines Chain of Thought (CoT) and Program-of-Thought \cite{chen2023program} rationales to teach LLMs to use external tools like Python interpreters for mathematical problem-solving.

To assess the correctness of reasoning steps, most existing work \cite{luo_wizardmath_2023, yu_metamath_2024} evaluates the quality by directly comparing the final answer. However, some early studies explore reasoning step quality differently. \cite{sawada_arb_2023} measures reasoning step quality by comparing the similarity between generated and reference reasoning. \cite{dubois_alpacafarm_2024} treats powerful LLMs as verifiers, asking them to generate judgments for the reasoning steps. \cite{xia2024evaluating} introduces a new methodology employing validity and redundancy to characterize reasoning quality, along with accompanying LLMs to assess them automatically.

Various methods extend LLMs as verifiers and demonstrate their usage for self-correction \cite{zhang2024small}. \cite{zheng_judging_2023} shows that models like GPT-4 align with human preferences, indicating their potential as tools for accessing LLM-generated responses. \cite{olausson_is_2024} finds that LLMs struggle to find their own reasoning errors in code generation but can correct them with adequate feedback. However, there's still a lack of clarity in math reasoning and using LLMs for mistake detection and rectification in foreign reasoning steps, not just their own self-generated reasoning steps. Our work focuses on LLMs' ability to correct MWPs reasoning steps and rectify them to reach the correct answer, as well as whether LLMs generalize to newer and complex datasets.

\vspace{-5pt}
\section{Conclusions}
\label{sec:conc}
\vspace{-5pt}
This study evaluates large language models (LLMs) like \gpto, \gpt4, \gptt, and smaller models (\llama, \mixtral, \PHI) on their ability to detect and correct errors in mathematical reasoning. Our \mwp dataset is meticulously curated with incorrect reasoning steps generated using both rule-based methods and smaller language models, ensuring a comprehensive evaluation of LLMs' error detection and correction capabilities. \gpto stands out, demonstrating superior performance in handling complex tasks and correcting mistakes. However, smaller models lag significantly, highlighting the need for advancements in their reasoning capabilities. The analysis also reveals concerns about data contamination and overfitting, particularly in \gpt's performance on GSM8K. A notable drop in performance on newer datasets like \mathb and \jee indicates challenges in generalizing to novel problems. Addressing these issues is crucial for improving LLMs' reliability and applicability in real-world mathematical problem-solving. Future research should focus on refining training processes, enhancing generalization, and mitigating data contamination to advance the field.
\newpage

\bibliography{main}
\bibliographystyle{abbrvnat}
\newpage
% \input{checklist}
% \newpage
\section*{Appendix}
The dataset and code to run all experiments will be made available soon.
%are provided in this \href{https://anonymous.4open.science/r/Exposing-the-Achille-Heel-1D11/}{repository}.
\section{\mwp Dataset}
\label{app:dataset}
\mwp dataset is curated using 4 different types of well-known datasets. Below are the details of each of the datasets.

\begin{itemize}
\itemsep0em 
    \item \gsm~\cite{cobbe_training_2021}:\gsm is a dataset of diverse grade school math word problems created by human writers, involving basic arithmetic operations. Released in November 2021.
    \item \mathd~\cite{hendrycks2021measuring}: The \mathd dataset is divided into seven categories, each with five difficulty levels. For our study, we used levels 1, 2, and 3 from the algebra and counting and probability categories. Released in November 2021.
    \item \mathb~\cite{liu2024mathbench}: \mathb is a recent dataset with questions divided by educational stages, from basic arithmetic to college levels. For our experiment, we chose middle and high-school-level single-choice multiple-choice questions. Released in May 2024.
    \item \jee~\cite{arora2023llms}: \jee is a challenging benchmark dataset for evaluating LLM problem-solving abilities, containing 515 pre-engineering math, physics, and chemistry problems from the IIT JEE-Advanced Exam. For our experiment, we chose mathematics single-choice questions. Released in October 2023.
\end{itemize}

\subsection{Prompts to curate reasoning steps in \mwp dataset}
\gsm and \mathd already contain MWP questions, a chain of thought reasoning steps and a final answer. To curate chain of thought reasoning step for \mathb and \jee we made use of GPT-4. While prompting GPT-4 we made sure that reasoning steps did not contain the final answer, so that final answer is not picked directly from the reasoning step. \autoref{lst:prompt1} prompt is used to curate the reasoning steps. 
\begin{minipage}{\linewidth}
\begin{lstlisting}[style=promptstyle,caption={Prompt to curate reasoning chain without answers.},label={lst:prompt1}]
Strictly follow the below conditions.
1. Output format: \nReasoning Chain: \nFinal Answer: 
2. Reasoning Chain should be separated by a new line only.
3. Reasoning chain cannot have the final answer. (Replace the final answer  in the reasoning chain with its calculation or ####)
4. Do not include any additional information in the final answer (only the answer).
\end{lstlisting}
\end{minipage}
Table~\ref{tab:rule_example} shows examples of default reasoning steps from \gsm dataset.
\begin{table}[h!]
\caption{Example of rule based incorrect reasoning step (\gsm dataset)}
\label{tab:rule_example}
\resizebox{\columnwidth}{!}{%
\begin{tabular}{|l|l|}
\hline
Question &
  \begin{tabular}[c]{@{}l@{}}Gerald spends \$100 a month on baseball supplies. \\ His season is 4 months long. \\ He wants to use the months he's not playing baseball \\ to save up by raking, shoveling, and mowing lawns. \\ He charges \$10 for each. How many chores does he need to average a month \\ to save up for his supplies?\end{tabular} \\ \hline
Final Answer &
  5 \\ \hline
Gold Reasoning step &
  \begin{tabular}[c]{@{}l@{}}He needs to save up \$400 because 4 x 100 = 400\\ He has 8 months to earn this money because 12 - 4 = 8\\ He needs to earn \$50 a month because 400 / 8 = 50\\ He needs to do 5 tasks a month because 50 / 10 = 5\end{tabular} \\ \hline
Shuffle reasoning step &
  \begin{tabular}[c]{@{}l@{}}He needs to earn \$50 a month because 400 / 8 = 50\\ He needs to save up \$400 because 4 x 100 = 400\\ He needs to do 5 tasks a month because 50 / 10 = 5\\ He has 8 months to earn this money because 12 - 4 = 8\end{tabular} \\ \hline
Delete reasoning step &
  \begin{tabular}[c]{@{}l@{}}He needs to save up \$400 because 4 x 100 = 400\\ He needs to earn \$50 a month because 400 / 8 = 50\\ He needs to do 5 tasks a month because 50 / 10 = 5\end{tabular} \\ \hline
Shuffle numerical values &
  \begin{tabular}[c]{@{}l@{}}He needs to save up \$400 because 4 x 100 = 400\\ He has 50 months to earn this money because \textcolor{red}{8 - 8 = 4}\\ He needs to earn \$12 a month because 400 / 8 = 50\\ He needs to do 5 tasks a month because 50 / \textcolor{red}{10} = 5\end{tabular} \\ \hline
Replace numerical values &
  \begin{tabular}[c]{@{}l@{}}He needs to save up \$400 because 4 x 100 = 400\\ He has 8 months to earn this money because 12 - 4 = 8\\ He needs to earn \textcolor{red}{\$6} a month because \textcolor{red}{32} / 8 = 50\\ He needs to do 76 tasks a month because 50 / 10 = 5\end{tabular} \\ \hline
Shuffle Operations &
  \begin{tabular}[c]{@{}l@{}}He needs to save up \$400 because 4 x 100 = 400\\ He has 8 months to earn this money because 12 \textcolor{red}{*} 4 = 8\\ He needs to earn \$50 a month because 400 \textcolor{red}{-} 8 = 50\\ He needs to do 5 tasks a month because 50 / 10 = 5\end{tabular} \\ \hline
Insert Random Reasoning step &
  \begin{tabular}[c]{@{}l@{}}He needs to save up \$400 because 4 x 100 = 400\\ \textcolor{red}{Therefore, Faye has \$60 - \$30 = \$30 left.}\\ He has 8 months to earn this money because 12 - 4 = 8\\ He needs to earn \$50 a month because 400 / 8 = 50\\ He needs to do 5 tasks a month because 50 / 10 = 5\end{tabular} \\ \hline
\end{tabular}%
}
\end{table}
\newpage
\section{SLMs reasoning steps}
\label{app:slm_reasoning_steps}
 SLMs were used to generate chain of thought (COT) reasoning step and final answers for all dataset questions. Each model \llama, \mixtral, \PHI where prompted using \autoref{lst:prompt1} to curate the reasoning step without an answer. If the final answer was incorrect we filtered out the reasoning steps as incorrect.
 
 Table~\ref{tab:slm_example} shows examples of SLM incorrect reasoning steps from \gsm dataset.

\begin{table}[h!]
\caption{Example of incorrect reasoning steps generated using SLM's (\gsm dataset)}
\label{tab:slm_example}
\resizebox{\columnwidth}{!}{%
\begin{tabular}{|l|l|}
\hline
Question &
  \begin{tabular}[c]{@{}l@{}}Gerald spends \$100 a month on baseball supplies. \\ His season is 4 months long. \\ He wants to use the months he's not playing baseball to save up by raking, shoveling, and mowing lawns. \\ He charges \$10 for each. How many chores does he need to average a month to save up for his supplies?\end{tabular} \\ \hline
Final Answer &
  5 \\ \hline
Gold Reasoning step &
  \begin{tabular}[c]{@{}l@{}}He needs to save up \$400 because 4 x 100 = 400.\\ He has 8 months to earn this money because 12 - 4 = 8\\ He needs to earn \$50 a month because 400 / 8 = 50\\ He needs to do 5 tasks a month because 50 / 10 = 5\\ \#\#\#\# 5\end{tabular} \\ \hline
\llama Incorrect reasoning step &
  \begin{tabular}[c]{@{}l@{}}Gerald spends \$100 on baseball supplies per month.\\ Season is 4 months long.\\ To save up for baseball supplies, Gerald needs to earn more than the cost of supplies per month.\\ Gerald charges \$10 for each chore he does.\\ To calculate the number of chores Gerald needs to do per month, \\ divide the cost of supplies by the amount he charges per chore.\end{tabular} \\ \hline
\llama Predicted Answer &
  "" \\ \hline
\mixtral Incorrect reasoning step &
  \begin{tabular}[c]{@{}l@{}}Gerald spends \$100 a month for 4 months, so his total spending for the season is 100*4 = \$400\\ Now, we need to find out how many chores he needs to do to earn this amount\\ He charges \$10 per chore, so we divide the total amount by the cost per chore: 400 / 10 = \#\#\#\#\end{tabular} \\ \hline
\mixtral Predicted Answer &
  40 \\ \hline
\PHI Incorrect reasoning step &
  \begin{tabular}[c]{@{}l@{}}Gerald's total expenditure for baseball supplies = \$100 per month * 4 months = \$400\\ Gerald earns \$10 per chore.\\ To find out how many chores he needs to do to save up \$400, \\ we divide his total expenditure by the amount he earns per chore.= \$400 / \$10= \#\#\#\#\end{tabular} \\ \hline
\PHI Predicted Answer &
  40 \\ \hline
\end{tabular}%
}
\end{table}
\section{Task T1 and T2}
\label{app:t1_t2_details}
Task T1 evaluates the model's ability to detect mistakes rectify them and derive the correct answer. \autoref{lst:prompt2} was used in a few shot settings for task T1. 
% \begin{minipage}{\linewidth}
\begin{lstlisting}[style=promptstyle,caption={Prompt for Task T1},label={lst:prompt2}]
You are a mathematics educator with a deep understanding of elementary and middle school mathematics. You are experienced in teaching multi-step problem-solving techniques and have a knack for breaking down complex problems into manageable steps. Your expertise lies in basic arithmetic operations such as addition, subtraction, multiplication, and division. You can provide clear, step-by-step solutions to mathematical problems that require multi-step reasoning. 

You are provided with a mathematical question and a step-by-step solution along with it. The solution might have some mistakes. Identify if the solution is correct or incorrect. If the solution is correct, output the final answer with the help of the solution provided. If the solution is incorrect, correct the existing solution and determine the final answer with the help of the corrected solution.
Reasoning chain Correct (Yes/No): 
Corrected reasoning chain or NA: 
Final answer (just the number):
\end{lstlisting}
% \end{minipage}
Task T2 evaulates the model's ability to detect mistake and solve MWP based on the provided reasoning step. \autoref{lst:prompt3} was used in a few shot setting for task T2. Here we insure that final answer is generated with the help of the reasoning steps provided, which may or may not be correct.

\begin{lstlisting}[style=promptstyle,caption={Prompt for Task T2},label={lst:prompt3}]
You are a mathematics educator with a deep understanding of elementary and middle school mathematics. You are experienced in teaching multi-step problem-solving techniques and have a knack for breaking down complex problems into manageable steps. Your expertise lies in basic arithmetic operations such as addition, subtraction, multiplication, and division. You can provide clear, step-by-step solutions to mathematical problems that require multi-step reasoning. 

You are provided with a mathematical question and a step-by-step solution along with it. The solution might have some mistakes. Identify if the solution is correct or incorrect and output the final answer based on the provided solution.
Reasoning chain Correct (Yes/No): 
Final answer (just the number):
\end{lstlisting}
\section{T2 Results}
Task T2 evaluates the performance in deriving the final answer based on the reasoning step which may or may not be correct. In task T2 we do not instruct the model to correct the reasoning step, and calcualate the final answer based on the provided reasoning step. Due to which we see a signifant drop in performance between Task T1 and Task T2. Table~\ref{tab:md_t2} presents the mistake detection performance (F1 score) of all the models with Task T2 and Table~\ref{tab:perf_t2} presents the performance in deriving the final answer (F1 Score) of all the models.
\label{app:t2_details}
\begin{table}[h!]
\centering
\caption{Mistake Detection Performance (F1 score) on \mwp dataset for Task T2. (D-Default reasoning steps, SM-Smaller model reasoning steps) (Bold: Best, Underline:Second best)}
\label{tab:md_t2}
\resizebox{0.8\columnwidth}{!}{%
\begin{tabular}{|l|rl|ll|ll|ll|lll|}
\hline
 &
  \multicolumn{2}{c|}{\gsm} &
  \multicolumn{2}{c|}{\mathd} &
  \multicolumn{2}{c|}{\mathb} &
  \multicolumn{2}{c|}{\jee} &
  \multicolumn{3}{c|}{Average} \\ \hline
Model &
  \multicolumn{1}{l}{D} &
  SM &
  D &
  SM &
  D &
  SM &
  D &
  SM &
  D &
  SM &
  Overall \\ \hline
\gpto &
  \multicolumn{1}{l}{\textbf{----}} &
  ---- &
  \textbf{----} &
  \textbf{----} &
  \textbf{----} &
  \textbf{----} &
  \textbf{----} &
  \textbf{----} &
  \textbf{----} &
  \textbf{----} &
  \textbf{----} \\
\gpt &
  0.67 &
  \multicolumn{1}{r|}{0.61} &
  \multicolumn{1}{r}{0.75} &
  0.76 &
  0.48 &
  0.88 &
  0.76 &
  0.85 &
  0.66 &
  0.78 &
  0.72 \\
\gptt &
  0.58 &
  \multicolumn{1}{r|}{0.40} &
  0.69 &
  0.42 &
  0.33 &
  0.24 &
  0.51 &
  0.41 &
  0.53 &
  0.36 &
  0.45 \\
\llama &
  0.11 &
  NA &
  0.22 &
  NA &
  0.11 &
  NA &
  0.75 &
  NA &
  0.30 &
  NA &
  0.30 \\
\mixtral &
  0.69 &
  NA &
  0.75 &
  NA &
  0.60 &
  NA &
  0.76 &
  NA &
  0.70 &
  NA &
  0.70 \\
\PHI &
  0.56 &
  NA &
  0.52 &
  NA &
  0.46 &
  NA &
  0.54 &
  NA &
  0.52 &
  NA &
  0.52 \\
\claude &
  \multicolumn{1}{l}{----} &
  \textbf{----} &
  ---- &
  ---- &
  ---- &
  ---- &
  ---- &
  ---- &
  ---- &
  ---- &
  ---- \\ \hline
\end{tabular}%
}
% \vspace{-12pt}
\end{table}

\begin{table}[h!]
\centering
\caption{Performance in deriving correct answers (F1 score) on \mwp dataset for Task T2. (D-Default reasoning steps, SM-Smaller model reasoning steps) (Bold: Best, Underline:Second best)}
\label{tab:perf_t2}
\resizebox{0.8\columnwidth}{!}{%
\begin{tabular}{|l|ll|ll|ll|ll|lll|}
\hline
 &
  \multicolumn{2}{l|}{\gsm} &
  \multicolumn{2}{l|}{\mathd} &
  \multicolumn{2}{l|}{\mathb} &
  \multicolumn{2}{l|}{\jee} &
  \multicolumn{3}{l|}{Average} \\ \hline
Model &
  \multicolumn{1}{l}{D} &
  SM &
  D &
  SM &
  \multicolumn{1}{l}{D} &
  SM &
  D &
  SM &
  D &
  SM &
  Overall \\ \hline
\gpto &
  \multicolumn{1}{l}{\textbf{----}} &
  ---- &
  \textbf{----} &
  \textbf{----} &
  \multicolumn{1}{l}{\textbf{----}} &
  \textbf{----} &
  \textbf{----} &
  \textbf{----} &
  \textbf{----} &
  \textbf{----} &
  \textbf{----} \\
\gpt &
  0.99 &
  \multicolumn{1}{l|}{0.65} &
  0.72 &
  0.48 &
  0.82 &
  \multicolumn{1}{l|}{0.27} &
  0.39 &
  0.29 &
  0.73 &
  0.42 &
  0.57 \\
\gptt &
  0.85 &
  \multicolumn{1}{l|}{0.26} &
  0.66 &
  0.31 &
  0.67 &
  \multicolumn{1}{l|}{0.16} &
  0.48 &
  0.20 &
  0.67 &
  0.23 &
  0.45 \\
\llama &
  0.84 &
  NA &
  0.33 &
  NA &
  0.44 &
  NA &
  0.36 &
  NA &
  0.49 &
  NA &
  0.49 \\
\mixtral &
  0.91 &
  NA &
  0.64 &
  NA &
  0.68 &
  NA &
  0.11 &
  NA &
  0.58 &
  NA &
  0.58 \\
\PHI &
  0.92 &
  NA &
  0.62 &
  NA &
  0.65 &
  NA &
  0.49 &
  NA &
  0.67 &
  NA &
  0.67 \\
\claude &
  \multicolumn{1}{l}{----} &
  \textbf{----} &
  ---- &
  ---- &
  \multicolumn{1}{l}{----} &
  ---- &
  ---- &
  ---- &
  ---- &
  ---- &
  ---- \\ \hline
\end{tabular}%
}
% \vspace{-12pt}
\end{table}
\section{Model Used}
Below are brief details of the models we have used for benchmarking our \mwp dataset.

\begin{enumerate}
    \item \textbf{\gpto:} \gpto is a multimodal model by OpenAI, and it has the same high intelligence as GPT-4 Turbo but is much more efficient—it generates text 2x faster and is 50\% cheaper. Additionally, GPT-4o has the best vision and performance across non-English languages of any OpenAI model. Last training data: October 2023. 
    \item \textbf{\gpt:} \gpt is a large multimodal model by OpenAI that can solve difficult problems with greater accuracy than any of OpenAI previous models, thanks to its broader general knowledge and advanced reasoning capabilities. Last training data: September 2021.
    \item \textbf{\gptt:} \gptt is a large language model by OpenAI GPT-3.5 that can understand and generate natural language or code and has been optimized for chat using the Chat Completions API but work well for non-chat tasks as well. Last training date: September 2021.
    \item \textbf{\claude:} \claude is Anthropic's most capable and intelligent model yet, ideal for navigating complex tasks like in-depth analysis, research, and task automation. Last training data: August 2023.
    \item \textbf{\llama:} Llama 2 is a collection of pretrained and fine-tuned generative text models ranging in scale from 7 billion to 70 billion parameters from meta. This is the 7B fine-tuned model, optimized for dialogue use cases. Training date: September 2022.
    \item \textbf{\mixtral:} Mixtral is a Mixture of Experts (MoE) model with 8 experts per MLP, with a total of 45 billion parameters. Despite the model having 45 billion parameters, the compute required for a single forward pass is the same as that of a 14 billion parameter model. This is because even though each of the experts have to be loaded in RAM (70B like ram requirement) each token from the hidden states are dispatched twice (top 2 routing) and thus the compute (the operation required at each forward computation) is just 2 X sequence\_length. 
    \item \textbf{\PHI:} The Phi-3-Mini-128K-Instruct is a 3.8 billion-parameter by microsoft, lightweight, state-of-the-art open model trained using the Phi-3 datasets. This dataset includes both synthetic data and filtered publicly available website data, with an emphasis on high-quality and reasoning-dense properties. Last training data: October 2023.
\end{enumerate}
\label{app:model_details}

\section{METEOR and BertScore results}
\label{app:meteor_results}
BertScore computes a similarity score for each token in the candidate sentence with each token in the reference sentence using the BERT embeddings. 
Metric for Evaluation of Translation with Explicit Ordering (METEOR) score is a metric that measures the quality of generated text based on the alignment between the generated text and the reference text. The metric is based on the harmonic mean of unigram precision and recall, with recall weighted higher than precision.

Table~\ref{tab:bert} and Table~\ref{tab:meteor} present the BertScore and Meteor Score respectively for all the datasets across all models. We observed that these two metric evaluations where not fully able to capture the nuance capabilities of LLMs in rectifying the mistakes within reasoning steps. This can be seen in the results. \gpto has a consistently high performance across all the dataset, but when you compare the BERTScore between the corrected reasoning step and ground truth reasoning step you see the rest of the models clearly performing better than \gpto. \gpt has performed better than \gptt in most datasets.

\begin{table}[h!]
\centering
\caption{{BERTscores for correct and incorrect final answers derived after mistake rectification across all models and datasets.}}
\label{tab:bert}
\resizebox{\columnwidth}{!}{%
\begin{tabular}{|l|l|ll|ll|ll|ll|ll|ll|}
\hline
Datasets &
  Models &
  \multicolumn{2}{l|}{\gpto} &
  \multicolumn{2}{l|}{\gpt} &
  \multicolumn{2}{l|}{\gptt} &
  \multicolumn{2}{l|}{\llama} &
  \multicolumn{2}{l|}{\mixtral} &
  \multicolumn{2}{l|}{\PHI} \\ \hline
 &
  \textbf{} &
  Correct &
  Incorrect &
  Correct &
  Incorrect &
  Correct &
  Incorrect &
  Correct &
  Incorrect &
  Correct &
  Incorrect &
  Correct &
  Incorrect \\ \hline
\multirow{2}{*}{\gsm}   & D  & 0.95 & 0.91 & 0.98 & 0.93 & 0.97 & 0.95 & 0.96 & 0.98 & 0.97 & 0.94 & 0.94 & 0.91 \\
                                       & SM & 0.83 & 0.82 & 0.84 & 0.82 & 0.84 & 0.82 & NA   & NA   & NA   & NA   & NA   & NA   \\ \hline
\multirow{2}{*}{\mathd} & D  & 0.88 & 0.90 & 0.96 & 0.93 & 0.95 & 0.93 & 0.96 & 0.88 & 0.95 & 0.92 & 0.90 & 0.87 \\
                                       & SM & 0.84 & 0.80 & 0.83 & 0.81 & 0.84 & 0.81 & NA   & NA   & NA   & NA   & NA   & NA   \\ \hline
\mathb                  & D  & 0.88 & 0.83 & 0.97 & 0.95 & 0.97 & 0.94 & 0.90 & 0.89 & 0.96 & 0.95 & 0.93 & 0.90 \\
                                       & SM & 0.82 & 0.82 & 0.85 & 0.82 & 0.84 & 0.83 & NA   & NA   & NA   & NA   & NA   & NA   \\ \hline
\jee                    & D  & 0.89 & 0.89 & 0.88 & 0.87 & 0.94 & 0.95 & 0.86 & 0.82 & 0.85 & 0.87 & 0.70 & 0.85 \\
                                       & SM & 0.86 & 0.87 & 0.85 & 0.86 & 0.78 & 0.86 & NA   & NA   & NA   & NA   & NA   & NA   \\ \hline
\end{tabular}%
}
% \vspace{-10pt}
\end{table}

\begin{table}[h!]
\centering
\caption{{Meteor Score for correct and incorrect final answers derived after mistake rectification across all models and datasets.}}
\label{tab:meteor}
\resizebox{\columnwidth}{!}{%
\begin{tabular}{|l|l|cc|cc|cc|cc|cc|cc|}
\hline
Datasets &
  Models &
  \multicolumn{2}{l|}{\gpto} &
  \multicolumn{2}{l|}{\gpt} &
  \multicolumn{2}{l|}{\gptt} &
  \multicolumn{2}{l|}{\llama} &
  \multicolumn{2}{l|}{\mixtral} &
  \multicolumn{2}{l|}{\PHI} \\ \hline
 &
  \textbf{} &
  \multicolumn{1}{l}{Correct} &
  \multicolumn{1}{l|}{Incorrect} &
  \multicolumn{1}{l}{Correct} &
  \multicolumn{1}{l|}{Incorrect} &
  \multicolumn{1}{l}{Correct} &
  \multicolumn{1}{l|}{Incorrect} &
  \multicolumn{1}{l}{Correct} &
  \multicolumn{1}{l|}{Incorrect} &
  \multicolumn{1}{l}{Correct} &
  \multicolumn{1}{l|}{Incorrect} &
  \multicolumn{1}{l}{Correct} &
  \multicolumn{1}{l|}{Incorrect} \\ \hline
\multirow{2}{*}{\gsm}   & D  & 0.81 & 0.54 & 0.92 & 0.62 & 0.88 & 0.77 & 0.87 & 0.83 & 0.85 & 0.74 & 0.77 & 0.66 \\
                                       & SM & 0.33 & 0.27 & 0.37 & 0.31 & 0.37 & 0.32 & NA   & NA   & NA   & NA   & NA   & NA   \\ \hline
\multirow{2}{*}{\mathd} & D  & 0.48 & 0.54 & 0.76 & 0.70 & 0.76 & 0.67 & 0.78 & 0.59 & 0.73 & 0.66 & 0.55 & 0.48 \\
                                       & SM & 0.32 & 0.28 & 0.30 & 0.26 & 0.33 & 0.28 & NA   & NA   & NA   & NA   & NA   & NA   \\ \hline
\mathb                  & D  & 0.55 & 0.35 & 0.82 & 0.63 & 0.82 & 0.68 & 0.49 & 0.57 & 0.81 & 0.68 & 0.67 & 0.53 \\
                                       & SM & 0.33 & 0.30 & 0.32 & 0.25 & 0.32 & 0.29 & NA   & NA   & NA   & NA   & NA   & NA   \\ \hline
\jee                    & D  & 0.37 & 0.31 & 0.30 & 0.22 & 0.49 & 0.54 & 0.15 & 0.13 & 0.53 & 0.46 & 0.20 & 0.25 \\
                                       & SM & 0.28 & 0.26 & 0.21 & 0.21 & 0.08 & 0.25 & NA   & NA   & NA   & NA   & NA   & NA   \\ \hline
\end{tabular}%
}
% \vspace{-10pt}
\end{table}

\section{Average reasoning Step Length}
\label{app:length_reasoning_step}
We noticed that the average word length of rectified reasoning for correct and incorrect for \gpto was higher than other models. Table~\ref{tab:avg_length} presents the average word length of the rectified reasoning step for all datasets across the models.
\begin{table}[h!]
\centering
\caption{{Average length of rectified reasoning steps on \mwp dataset}}
\label{tab:avg_length}
\resizebox{0.8\columnwidth}{!}{%
\begin{tabular}{|l|cc|cc|cc|cc|ccc|}
\hline
 &
  \multicolumn{2}{l|}{\gsm} &
  \multicolumn{2}{l|}{\mathd} &
  \multicolumn{2}{l|}{\mathb} &
  \multicolumn{2}{l|}{\jee} &
  \multicolumn{3}{l|}{Average} \\ \hline
Model &
  \multicolumn{1}{l}{D} &
  \multicolumn{1}{l|}{SM} &
  \multicolumn{1}{l}{D} &
  \multicolumn{1}{l|}{SM} &
  \multicolumn{1}{l}{D} &
  \multicolumn{1}{l|}{SM} &
  \multicolumn{1}{l}{D} &
  \multicolumn{1}{l|}{SM} &
  \multicolumn{1}{l}{D} &
  \multicolumn{1}{l}{SM} &
  \multicolumn{1}{l|}{Overall} \\ \hline
\gpto    & 100.14 & 131.47 & 147.50 & 182.69 & 312.11 & 323.45 & 647.66 & 619.09 & 301.85 & 314.18 & 308.01 \\
\gpt     & 66.59  & 122.24 & 79.32  & 121.59 & 146.54 & 140.43 & 356.71 & 322.53 & 162.29 & 176.69 & 169.49 \\
\gptt    & 66.58  & 126.30 & 94.17  & 124.56 & 140.50 & 177.36 & 670.34 & 338.53 & 242.90 & 191.69 & 217.29 \\
\llama   & 44.73  & NA     & 113.35 & NA     & 177.67 & NA     & 137.05 & NA     & 118.20 & NA     & 118.20 \\
\mixtral & 63.04  & NA     & 88.26  & NA     & 140.57 & NA     & 402.79 & NA     & 173.67 & NA     & 173.67 \\
\PHI     & 84.92  & NA     & 115.10 & NA     & 172.57 & NA     & 293.90 & NA     & 166.62 & NA     & 166.62 \\
\claude  & 62.18  & 138.91 & 70.60  & 134.05 & 144.85 & 192.84 & 561.88 & 438.44 & 209.88 & 226.06 & 217.97 \\ \hline
\end{tabular}%
}
% \vspace{-12pt}
\end{table}
\section{Data contamination and Memorization}
\label{app:memo_prompt}
Two different types of prompts were used to test out memorization of individual instances of the dataset. The reasoning step was split into two halves, and models were prompted in two different ways (general and guided) to complete the second half of the reasoning step. To detect contamination we evaluated the overlap score between generated and reference instances using ROUGE-L Score. \autoref{lst:prompt4} and \autoref{lst:prompt5} shows the prompt used for guided and generation instructions respectively.
\begin{lstlisting}[style=promptstyle,caption={Guided instruction for dataset GSM8K},label={lst:prompt4}]
You are provided with a full question and the first answer piece from the test split of the GSM8K dataset. Finish the second answer piece of the instance as exactly appeared in the dataset. Only rely on the original form of the instance in the dataset to finish the second answer piece. Output only the second answer piece.
\end{lstlisting}
\begin{lstlisting}[style=promptstyle,caption={General instruction for dataset GSM8K},label={lst:prompt5}]
Based on the provided question, finish the second answer piece based on the first answer piece, such that these two pieces become a single instance answer. Output only the second answer piece.
\end{lstlisting}
Here \gsm and test are the extra information provided for the model to uniquely identify instances from the source dataset and complete the reasoning step.

Table~\ref{tab:mem} presents the complete result for the average ROUGE-L score of guided and general for all datasets across all models.
\begin{table}[h!]
\centering
\caption{{Rouge L score between guided and general instructions on \mwp dataset}}
\label{tab:mem}
\resizebox{\columnwidth}{!}{%
\begin{tabular}{|l|l|ll|ll|ll|ll|ll|ll|}
\hline
Datasets &
  Models &
  \multicolumn{2}{l|}{\gpto} &
  \multicolumn{2}{l|}{\gpt} &
  \multicolumn{2}{l|}{\gptt} &
  \multicolumn{2}{l|}{\llama} &
  \multicolumn{2}{l|}{\mixtral} &
  \multicolumn{2}{l|}{\PHI} \\ \hline
 &
  \textbf{} &
  Guided &
  General &
  Guided &
  General &
  Guided &
  General &
  Guided &
  General &
  Guided &
  General &
  Guided &
  General \\ \hline
\multirow{2}{*}{\gsm} &
  D &
  0.57 &
  0.44 &
  \textbf{0.67} &
  0.56 &
  \textbf{0.53} & 0.49 & 0.26 & 0.28 & 0.46 & 0.44 & 0.32 & 0.32 \\
 & SM &0.55 & 0.51 & 0.57 & 0.55 &
  0.49 &
  0.47 &
  0.30 &
  0.32 &
  0.55 &
  0.50 &
  0.42 &
  0.41 \\ \hline
\multirow{2}{*}{\mathd} &
  D &
  0.44 &
  0.25 &
  0.52 &
  0.48 &
  0.39 &
  0.38 &
  0.25 &
  0.26 &
  0.39 &
  0.32 &
  0.26 &
  0.27 \\
 &
  SM &
  0.51 &
  0.38 &
  0.54 &
  0.54 &
  0.45 &
  0.44 &
  0.30 &
  0.29 &
  0.48 &
  0.46 &
  0.38 &
  0.39 \\ \hline
\mathb &
  D &
  0.43 &
  0.41 &
  0.48 &
  0.46 &
  0.38 &
  0.36 &
  0.26 &
  0.28 &
  0.36 &
  0.36 &
  0.30 &
  0.30 \\
 &
  SM &
  0.40 &
  0.38 &
  0.43 &
  0.42 &
  0.39 &
  0.38 &
  0.30 &
  0.33 &
  0.40 &
  0.38 &
  0.29 &
  0.30 \\ \hline
\jee &
  D &
  0.43 &
  0.39 &
  0.42 &
  0.40 &
  0.34 &
  0.33 &
  0.27 &
  0.25 &
  0.38 &
  0.34 &
  0.33 &
  0.31 \\
 &
  SM &
  0.32 &
  0.29 &
  0.34 &
  0.35 &
  0.31 &
  0.24 &
  0.22 &
  0.25 &
  0.26 &
  0.27 &
  0.20 &
  0.22 \\ \hline
\end{tabular}%
}
% \vspace{-10pt}
\end{table}

\section{Running Experiment Multiple Times}
\label{app:error_bar}

While running experiments on all models (LLMs and SLMs) we used the default hyperparameters to generate tokens. We ran a subset of the dataset on different prompt variations and saw comparable performance for various prompts. Due to the limitation of the API key, we were only able to run \gpto model on the \gsm dataset. On rerun we got very similar results, with an error rate of <= 0.01. 

% \section{Output from each model}
% \label{app:output_msp}
% The raw output of each model has been provided in this \href{https://anonymous.4open.science/r/Exposing-the-Achille-Heel-1D11/}{repository}. Additional details are present in the README.md file of the repository.

\end{document}